%% file: main.tex
\documentclass[10pt,twocolumn,letterpaper]{article}

\usepackage[pagenumbers]{cvpr} 

\usepackage{capt-of}
\usepackage[dvipsnames]{xcolor}
\usepackage{xspace}
\usepackage{enumitem}
\usepackage{amsmath,amssymb,amsbsy,amsfonts,dsfont,pifont,bm,bbm,mathrsfs,mathtools,nicefrac}
\usepackage{algorithm,algpseudocode,listings}
\usepackage{booktabs,multirow,adjustbox,diagbox,threeparttable,tabularray}
\usepackage{colortbl}
\usepackage{makecell}
\usepackage{marvosym}

\usepackage{tcolorbox}
\usepackage{etoc}
\usepackage{titletoc}

\newcommand{\var}{\texttt} 
\newcommand{\VarSty}[1]{\textnormal{\ttfamily\color{blue!90!black}#1}\unskip} 

\input{math_commands.tex}

\newcommand{\tocite}[1]{{\color{red} [TO CITE]}}

\definecolor{cvprblue}{rgb}{0.21,0.49,0.74}
\usepackage[pagebackref,breaklinks,colorlinks,citecolor=cvprblue]{hyperref}
\usepackage{wrapfig}
\usepackage[capitalize]{cleveref}  
\crefname{section}{Sec.}{Secs.}
\Crefname{section}{Section}{Sections}
\crefname{table}{Tab.}{Tabs.}
\Crefname{table}{Table}{Tables}
\crefname{figure}{Fig.}{Figs.}
\Crefname{figure}{Figure}{Figures}
\crefname{equation}{Eq.}{Eqs.}
\Crefname{equation}{Equation}{Equations}
\def\paperID{}
\def\confName{CVPR}
\def\confYear{2024}

\newcommand\nonumfootnote[1]{%
\begingroup%
    \renewcommand\thefootnote{}\footnote{\hspace{-3.7pt}#1}%
    \addtocounter{footnote}{-1}%
\endgroup%
}
             
\title{TagAlign: Improving Vision-Language Alignment with Multi-Tag Classification}

\author{
        Qinying Liu\textsuperscript{1,*,$\dagger$}, ~~
        Wei Wu\textsuperscript{1,*,$\dagger$}, ~~
        Kecheng Zheng\textsuperscript{2,3,\Letter}, ~~
        Zhan Tong\textsuperscript{3}, ~~ 
        Jiawei Liu\textsuperscript{1}, \\
        Yu Liu\textsuperscript{4}, ~~
        Wei Chen\textsuperscript{2}, ~~
        Zilei Wang\textsuperscript{1,\Letter}, ~~
        Yujun Shen\textsuperscript{3}
        \\[6pt]
        $^1$USTC ~~
        $^2$Zhejiang University ~~
        $^3$Ant Group ~~
        $^4$Alibaba Group\\[8pt]
        }

\begin{document}

\maketitle

\iftrue
\nonumfootnote{$^*$ Equal contribution
~
\Letter\ Corresponding authors
~
$\dagger$ Interns at Ant Group
}
\fi

\input{sections/0.abs.tex}
\input{sections/1.intro.tex}

\input{sections/2.related_work.tex}
\input{sections/3.method.tex}
\input{sections/4.exp.tex}
\input{sections/5.conclusion.tex}
\input{sections/6.ref.tex}

\input{sections/7.appendix.tex}

\end{document}

%% file: math_commands.tex
\usepackage{amsmath,amsfonts,bm}

\def\eqref#1{equation~\ref{#1}}

\def\1{\bm{1}}

\def\rmM{{\mathbf{M}}}

\def\rmZ{{\mathbf{Z}}}

\def\vc{{\bm{c}}}

\def\vp{{\bm{p}}}

\def\vt{{\bm{t}}}

\def\vw{{\bm{w}}}
\def\vx{{\bm{x}}}
\def\vy{{\bm{y}}}
\def\vz{{\bm{z}}}

\DeclareMathAlphabet{\mathsfit}{\encodingdefault}{\sfdefault}{m}{sl}
\SetMathAlphabet{\mathsfit}{bold}{\encodingdefault}{\sfdefault}{bx}{n}



%% file: sections/0.abs.tex
\begin{abstract}

The crux of learning vision-language models is to extract semantically aligned information from visual and linguistic data.
Existing attempts usually face the problem of coarse alignment, \textit{e.g.}, the vision encoder struggles in localizing an attribute-specified object.
In this work, we propose an embarrassingly simple approach to better align image and text features with no need of additional data formats other than image-text pairs.
Concretely, given an image and its paired text, we manage to parse objects (\textit{e.g.}, cat) and attributes (\textit{e.g.}, black) from the description, which are highly likely to exist in the image.
It is noteworthy that the parsing pipeline is fully automatic and thus enjoys good scalability.
With these parsed semantics as supervision signals, we can complement the commonly used image-text contrastive loss with the multi-tag classification loss.
Extensive experimental results on a broad suite of semantic segmentation datasets substantiate the average 5.2\% improvement of our framework over existing alternatives.
Furthermore, the visualization results indicate that attribute supervision makes vision-language models accurately localize attribute-specified objects.
Project page can be found at \href{https://qinying-liu.github.io/Tag-Align}{https://qinying-liu.github.io/Tag-Align}.

\end{abstract}

%% file: sections/1.intro.tex
\section{Introduction}\label{sec:intro}

\begin{figure}[t]
	\centering
        \includegraphics[width=1.\linewidth]{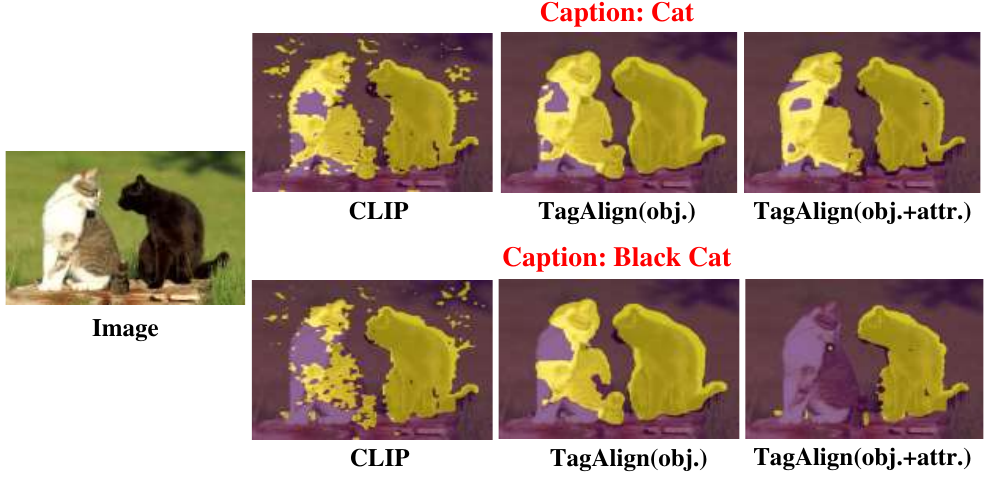}

	\caption{
        \textbf{Illustration of the effect of various tag supervisions (\textit{e.g.}, object and attribute) for open-vocabulary semantic segmentation.} The vanilla CLIP struggles in localizing an attribute-specified object. When introducing object supervision (as depicted in the third column), CLIP can focus on the more accurate region of text-specified object (\textit{e.g.}, cat). In addition, adding attribute supervision brings CLIP a stronger understanding of visual attribute-related concepts (as depicted in the fourth column). Best viewed in color.
        } 
	\label{fig:motivation}
\end{figure}

\begin{figure}[t]
	\centering
        \includegraphics[width=1.\linewidth]{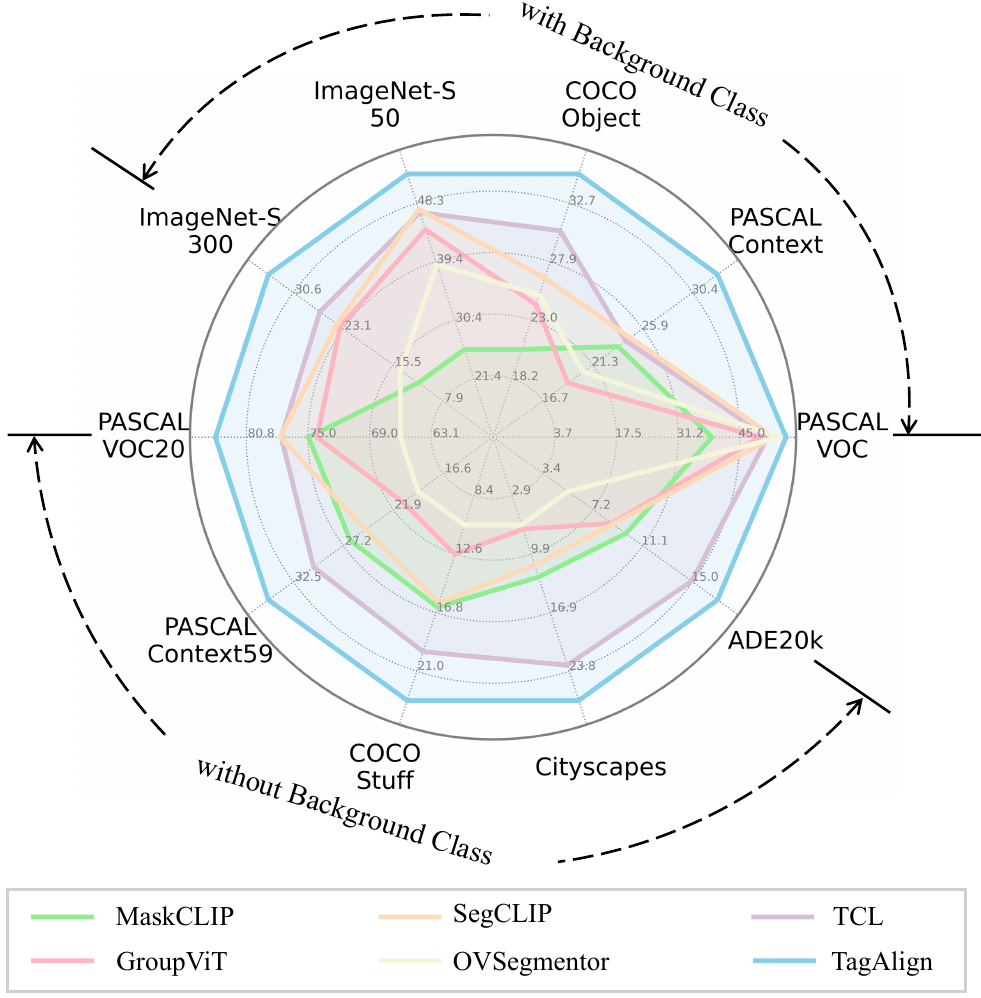}

	\caption{
        \textbf{Performance Comparison on various semantic segmentation benchmarks.}
        TagAlign outperforms existing methods by a large margin on all these benchmarks. Best viewed in color.
        } 
	\label{fig:radar}
\end{figure}

Key aspect in the process of learning vision-language (V\&L) encoders, \textit{e.g.}, CLIP~\cite{radford2021learning}, is to effectively extract semantically aligned information from both visual and linguistic data.
A well-aligned vision-language encoder stands out for its remarkable versatility, which includes unprecedented quality in text-to-image generation, as well as the improvement in multi-modality understanding tasks.
However, existing attempts usually face the problem of coarse alignment, \textit{e.g.}, the vision encoder struggles in localizing an attribute-specified object (as depicted in~\cref{fig:motivation}), making it hard to achieve precise localization of arbitrary concepts.

To improve the precise multi-modality alignment, various methods draw some supports from large-scale supervised data with manual annotations.
Specifically, some supervised data (\textit{i.e.}, mask, bounding box, or category) such as SA-1B~\cite{kirillov2023segment}, OpenImages~\cite{kuznetsova2020open}, and JFT300M~\cite{sun2017revisiting} has been proven to be critical for improving precise localization of arbitrary categories.
Despite the performance improvements observed in commonly used categories, the utilization of such annotated datasets necessitates rigorous efforts in data gathering, sampling, and human annotation, limiting their scalability and generalization.

Thereupon, recent researchers have turned their attention to how to automatically parse the data of image-text pairs, especially text.
Concretely, automatic text parsing tool (\textit{i.e.}, NLTK) has become a \textit{de-facto} approach in V\&L field, which can generate some tags from caption of image.
While these tags can bring performance improvements in V\&L models, these endeavors, incorporating much noise from NLTK, may restrict the models' ability to reach the upper limit of accurately perceiving semantic concepts.
For collecting higher-quality data, Tag2Text~\cite{huang2023tag2text} designs a universal and unified label system with an efficient data annotation engine, which generate about $4,000$ tags from some image-text datasets (\textit{e.g.,} CC12M).
Based on these tags, the series works of RAM~\cite{huang2023tag2text,ram} explore an image-tag recognition system with \textit{decoder module}, which can classify commonly human-used categories.
Additionally, DiHT~\cite{radenovic2023filtering} distills the knowledge from an extra multi-tag classification model, trained on collected private tags, only into the \textit{visual encoder} of CLIP. 
These methods focus on improving the classification performance of visual encoders, without explicitly exploring the alignment between vision and language.
A core viewpoint still worth exploring: \textit{how to utilize various multi-tags to help alignment between the image and text embeddings from V$\&$L encoders?}
I argue that there are two important questions in the training design of multi-modal encoders based on multi-tags.
The first one is the impact about the number of tag. Even using high-end tag signals in prior methods, a limited number of tags makes them fail to fully unleash the potential of multiple tags. 
Furthermore, the second one is that there is a lack of comprehensive exploration on how to effectively utilize various tags (\textit{e.g.,} attribute and objects).

In this paper, we propose an embarrassingly simple approach to improve the alignment between image and text features without the need for additional data formats beyond image-text pairs. 
Specifically, given an image and its paired text, we automatically parse objects (\textit{e.g.}, cat) and attributes (\textit{e.g.}, brown) from the description, which are highly likely to exist in the image. 
This parsing pipeline is implemented by a large language model, which is high-end and fully automatic, ensuring quality and scalability.
By using these parsed semantics as supervision signals, we can complement the commonly used image-text contrastive loss with a multi-tag classification loss, enabling the model to better capture the visual semantic concepts referred to in the texts.
Through that, our method becomes capable of accurately localizing text-specified objects (\textit{e.g.}, attribute-specified object), fulfilling more precise alignment between image and text.
Extensive experimental results as shown in~\cref{fig:radar} on a broad suite of semantic segmentation datasets (10 datasets) substantiate the average 5.2\% improvement of our framework over existing alternatives.
Moreover, it is mentioned that we only use the CC12M dataset with the parsed text as the training set, indicating that our method is both efficient and effective.

Besides remarkable performance on open-vocabulary semantic segmentation tasks, TaskAlign also enjoys the high flexibility to incorporate attribute information.
Concretely, the visualization results, as shown in~\cref{fig:motivation}, show that the inclusion of attribute supervision enables CLIP to accurately localize the attribute-specified object (`brown cat'). 
This finding indicates that different types of tag supervision have the potential to enhance the ability of cross-modal alignment in different aspects of CLIP.

%% file: sections/2.related_work.tex
\section{Related Work}\label{sec:related}
\noindent\textbf{Image-text alignment} is the key of vision-language models (VLMs) . 
Recent VLMs such as CLIP~\cite{radford2021learning} and Align~\cite{jia2021scaling} jointly train vision encoder and language encoder through contrastive learning.
However, current commonly used contrastive learning only act on the cls token from image encoder and the end token from text encoder, which leads to coarse alignment between visual and linguistic data. 
To address this issue, previous approaches attempt to improve image-text alignment by incorporating additional annotated data to supervise the training process.
GLIP~\cite{li2022grounded} trains the image encoder and text encoder to predict correct paired detected bounding boxes and phrases in the given text prompt.
SAM-CLIP~\cite{wang2023sam} merges CLIP and Segment Anything Model into a unified model by performing multi-task distillation on large scale image-text paired dataset and SA-1B~\cite{kirillov2023segment}.
The reliance of manual annotated dataset poses limitations on the scalability and generalization of vision-language models.

\noindent\textbf{Language-Supervised Semantic Segmentation} aims to learn segmentation from only image-text pairs without any dense annotation.
Existing methods can be categorized into two groups. The first group modifies or fine-tunes pre-trained CLIP models for segmentation.
MaskCLIP~\cite{ding2023maskclip} modifies the last attention layer of the CLIP image encoder to obtain a dense image embedding for segmentation. 
SimSeg~\cite{yi2023simple} sparsely samples a portion of patches and words used for the bidirectional contrastive losses. 
TCL~\cite{cha2023learning} introduces a region-level text grounder to produce text-grounded masks, then performs matching on grounded image regions and texts. 
The second group focuses on designing effective visual encoders for segmentation and further trains the model using image-text pairs.     
GroupViT~\cite{xu2022groupvit} proposes to group patch tokens into arbitrary-shaped segments with learnable group tokens in a bottom-up manner.
ViL-Seg~\cite{liu2022open} uses an online clustering head trained via mutual information maximization to group and classify segments. SegCLIP~\cite{luo2023segclip} adds a reconstruction loss and a superpixel-based KL loss to the normal image-text contrastive loss. OVSegmentor~\cite{xu2023learning} proposes a slot attention-based binding module to group patch tokens and then aligns the image embedding with the text embedding via an image-text contrast loss and a cross-image mask consistency loss.

\noindent\textbf{Multi-Label Classification}, also known as multi-tag recognition~\cite{wu2023mofi,singh2022revisiting,mahajan2018exploring,joulin2016learning}, is a fundamental computer vision task that aims to identify multiple tags for a given image. 
However, most existing multi-label datasets rely on manual annotations, which are labor-intensive and difficult to scale up. 
To address this issue, RAM related works~\cite{ram,huang2023tag2text,ram+} leverage text semantic parsing technical to efficiently obtain image tags and create a large-scale image tagging dataset consisting of about 4,000 commonly used categories. 
Meanwhile, private multi-tag data, named Instagram hashtags (IG-3.6B), are proposed in weakly supervised pretraining~\cite{singh2022revisiting,singh2022revisiting,mahajan2018exploring,joulin2016learning}.
Thanks to the high-end multi-tags, it can help the visual encoders to achieve superior tag recognition capabilities.
Although the recent RAM-related works share some similarities with our TagAlign in multi-tag classification, there are several fundamental differences.
Specifically, they introduce the decoder fusion module to process the image and text embeddings together, which not directly consider the embedding alignment between language and image.
We argue that improving the cross-modality embedding alignment of encoders from metric learning perspective~(\textit{e.g.}, dataset-wise multi-tag classification loss) is a key aspect in vision-language field.
Then, our method has analyzed the effects of attributes and objects separately, which may bring more insights about how to intelligently utilize the different kinds of tags.
Furthermore, the number of tag should be explored, which may be highly related to ability of open-vocabulary understanding.

%% file: sections/3.method.tex
\begin{figure*}[t]
	\centering
	\includegraphics[width=0.99\linewidth]{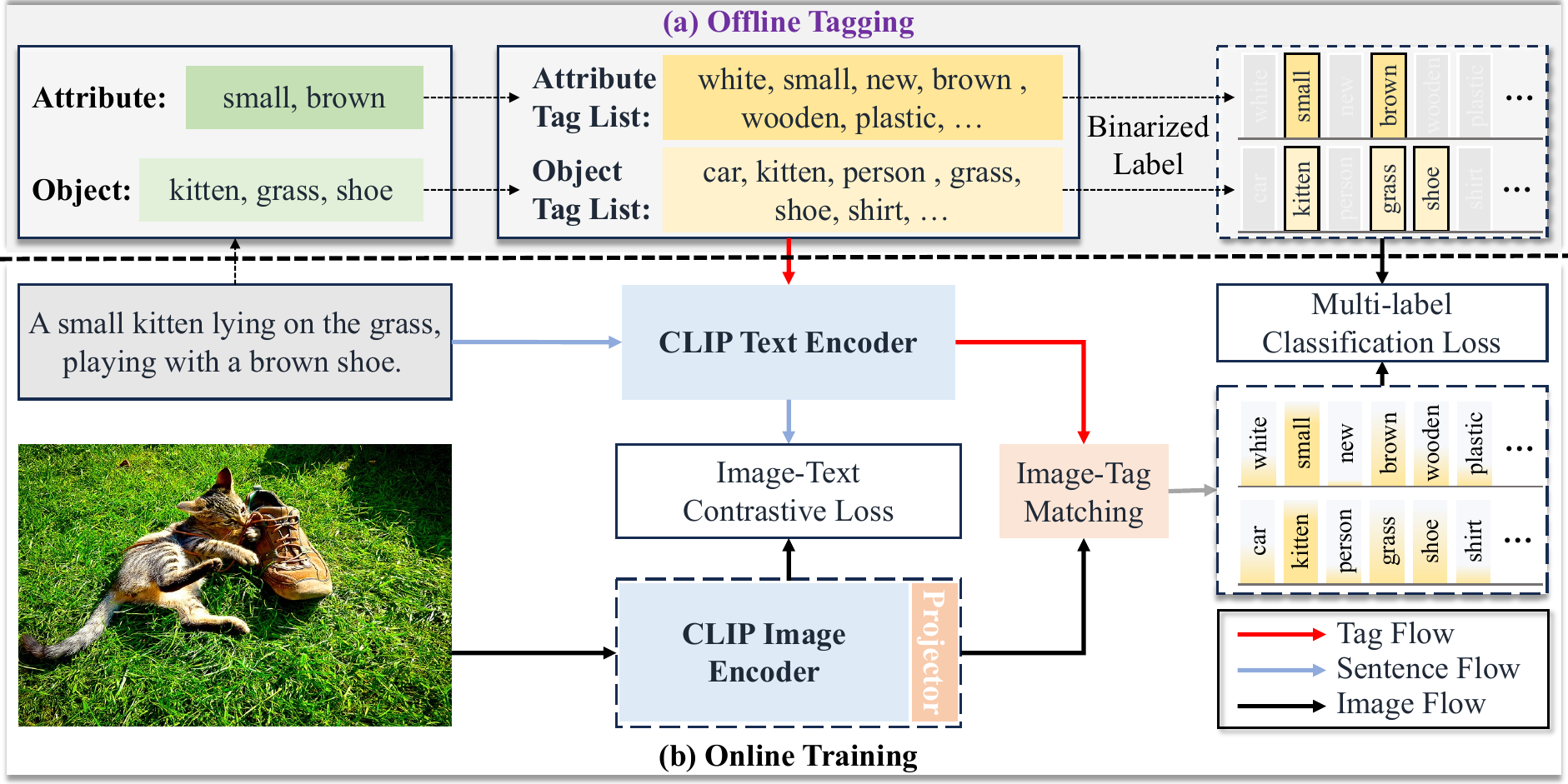}
	\centering
 \caption{\textbf{The framework of TagAlign.} At the core of TagAlign are two key components: a) LLM-aided tag parsing that parses the image captions into diverse tags (\ie, objects and attributes); b) Multi-tag classification that utilizes the parsed tags to supervise the model training. Best viewed in color.} 
	\label{fig:framework}
\end{figure*}

\section{Method}\label{sec:method}

Our method comprises two key ingredients: 1) LLM-aided tag parsing, which aims to extract diverse tags (\eg, objects and attributes) from the captions of the images; 2) multi-tag classification, which targets to guide the model training via a multi-tag classification objective.  
To be specific, given an image-caption pair, we prompt LLM to analyze the caption and extract tags from it. As a result, a large tag list can be established from the whole image-caption dataset. By indexing on this tag list, we transform the tags of each image to a binarized label, which enables us to use the label to supervise the model training with a simple multi-tag classification loss.  

\subsection{LLM-Aided Tag Parsing}  
In this section, we provide a detailed explanation of our proposed method for extracting tags from captions. Traditional methods~\cite{xu2022groupvit, xu2023learning} commonly employ the Natural Language ToolKit (NLTK)~\cite{loper2002nltk} for text parsing. However, as illustrated in~\cref{fig:compare_tags},  NLTK analyzes the texts solely based on Part-of-Speech (POS), leading to challenges in understanding the scene described in the caption.
On the other hand, large language models (LLMs) have demonstrated exceptional instruction-following capabilities, rendering them more controllable and flexible when parsing the texts. In light of this, we propose to prompt existing LLM to analyze and extract tags (\ie, objects and attributes) from given captions. 
Without loss of generality, we focus on objects for illustration purposes in the following.

\noindent\textbf{Prompt Engineering.} 
The prompts we employ consist of specific~\textit{rules} and~\textit{examples}. Firstly, a set of rules is crafted to constrain the LLM's responses, ensuring a focus on tangible and visible objects likely to exist in the image. Secondly, examples are provided to guide the LLM in generating desired response format. These examples leverage the in-context learning ability of LLM to instruct it on how to structure and format the output accurately. The Vicuna-33b~\cite{chiang2023vicuna} is used as the LLM to extract tags in our implementation. 
We refer to Appendix for more details about the used prompts. 

With the prompts in place, we query LLM to extract tags from the captions.  In~\cref{fig:compare_tags}, we offer examples to illustrate the comparison between tags extracted by LLM and NLTK.  As demonstrated, LLM offers two significant advantages over NLTK.
Firstly, it demonstrates a more accurate understanding of the scene described in the caption, resulting in a more precise listing of tangible and visible objects. Secondly, LLM performs word splitting in a more logical manner, mitigating the issue of segmenting entire objects into multiple components.

\begin{figure*}[t]
    \centering
    \includegraphics[width=0.98\linewidth]{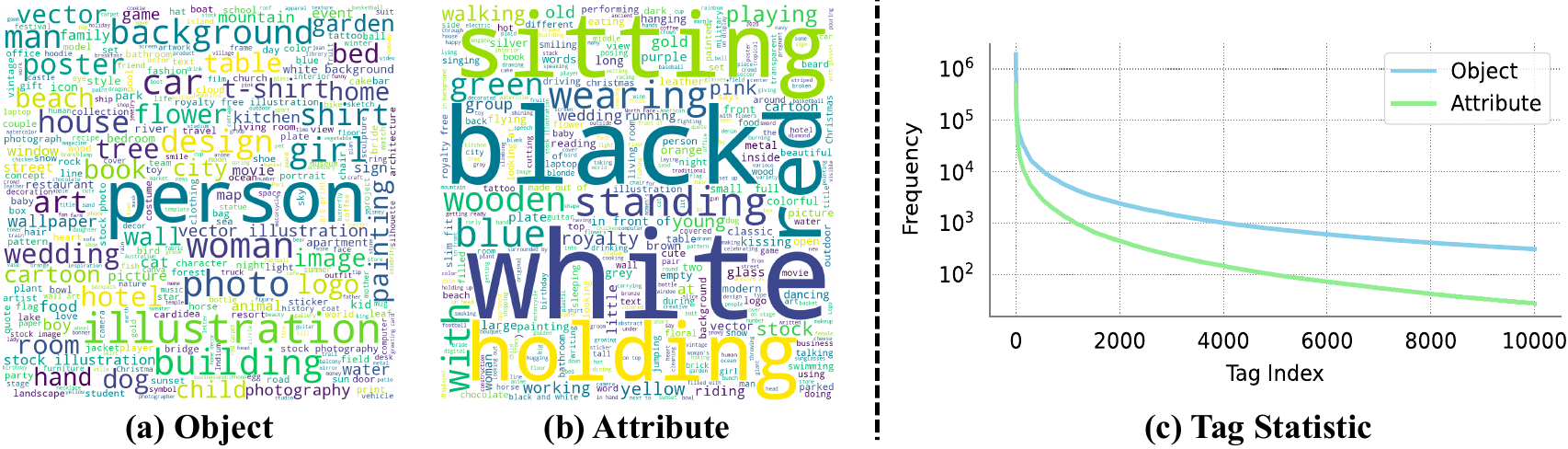}
    \caption{
    \textbf{Illustration of the most frequent tags in the (a) object tag list and  (b) attribute tag list.} The font size corresponds to the tag's frequency of occurrence in the tag list: the more frequent tag uses a larger font size.
    \textbf{(c) Statistic of object \& attribute tags from LLM parser}. The distribution of the tags exhibits a long-tailed pattern. Best viewed in color.}
    \label{fig:tags}
\end{figure*}

\noindent\textbf{Tag List Construction.}
Given an image-caption dataset consisting of image-caption pairs, each caption is sequentially processed by the LLM to extract tags. Subsequently, a comprehensive tag list is generated by aggregating all extracted tags and eliminating duplicates. 
To ease the pre-training process, we selectively retain only the top $K$ tags that appear most frequently. The most frequent object and attribute tags extracted by LLM are illustrated in~\cref{fig:tags}(a)(b).

\noindent\textbf{Binarized Label Construction.} After establishing the tag list, we proceed to construct a binarized label $\mathbf{y} \in \mathbb{R}^K$ for each image caption based on the presence of the extracted tags.
Specifically, for each element $\mathbf{y}_k$ in the label vector $\mathbf{y}$, we set it to 1 if the $k$-th tag in the tag list exists in the extracted tags from the caption. Conversely, if the $k$-th tag does not appear in the extracted tags, we assign $\mathbf{y}_k$ a value of 0. 

\subsection{Multi-Tag Classification}
In this section, we elaborate on how to leverage the binarized labels to guide the model training, thereby boosting the fine-grained image-text alignment.  

As shown in~\cref{fig:framework}, for an image $\vx$, we first pass it through the pre-trained image encoder of CLIP to obtain the patch embeddings (excluding the~\texttt{[CLS]} embedding). These patch embeddings are then processed by a lightweight projector, resulting in the final patch feature map denoted as $\rmZ \in \mathbb{R}^{L \times D}$, where $L$ is the number of patches. 
With these patch-level embeddings, our objective is to achieve fine-grained image-text alignment, despite the availability of only image-level labels for training.
This is a special case of Multiple Instance Learning (MIL), in which a ``bag'' (image) contains multiple ``instances'' (patches), and the labels of these instances collectively determine the label of the bag. 
Following the standard practice in MIL~\cite{wang2019comparison}, it is necessary to train a classifier for predicting the bag-level labels (\ie, image-level tags).  Once trained, the  classifier can be applied to the instances (patches), where the instances triggering the bag label should exhibit high activations.

To implement this approach, we first apply average pooling over the patch feature map to obtain the image-level representation $\vz \in \mathbb{R}^{D}$, calculated as $\vz = \frac{1}{L}\sum_{l=1}^{L} \rmZ_l$. 
Subsequently, $\vz$ is fed into a tag classifier to predict tag classification scores. To leverage the zero-shot transfer capability of CLIP, instead of using randomly initialized weights, we leverage the pre-trained text encoder from CLIP to generate~\textit{fixed} weights for the $K$ tags. More specifically, the tags are firstly prompted with the template ``A photo of a \{tag\}'', and then the prompted texts are processed by the text encoder, resulting in tag embeddings, denoted by $\{\vc_1, .., \vc_K\}$. These tag embeddings serve as the weights for the tag classifier. As a result, the tag classification scores can be computed as the relative similarity between the image-level representation and the tag embeddings.
Formally, the $k$-th tag classification score $\vp_k$ is expressed as:
\begin{equation}
\begin{aligned}
\vp_k = \frac{
\exp(\rho \cdot \operatorname{cos}({\vz}, \vc_k) )
}
{\sum_{i=1}^{K} \exp(\rho \cdot \operatorname{cos}({\vz}, \vc_i))
}.
\label{eq:cls_score}
\end{aligned}
\end{equation}
where $\rho$ is a learnable temperature scalar. Finally, we optimize the model through a multi-label cross-entropy loss between the tag classification scores $\vp$ and the binarized tag labels $\vy$: 
\begin{equation}
\begin{aligned}
\mathcal{L}_{\mathrm{Tag}} = - \frac{1}{|\vy|} \sum_{k=1}^{K}  \vy_k \log \vp_k.
\label{eq:cls_loss}
\end{aligned}
\end{equation}
Here, $|\vy|$ represents the number of positive labels in $\vy$.  

\noindent\textbf{Understanding from the Image-text Alignment Perspective.}
To enable a better understanding of TagAlign on why it works well, here, we analyze~\cref{eq:cls_loss} from the perspective of image-text alignment. 

First, we rewrite~\cref{eq:cls_score} as:
\begin{equation}
\begin{aligned}
\vp_k = \frac{
\exp(\rho \cdot \operatorname{cos}({\vz}, \vc_k) )
}
{\exp(\rho \cdot \operatorname{cos}({\vz}, \vc_k) ) + \sum_{i \neq k} \exp(\rho \cdot \operatorname{cos}({\vz}, \vc_i))
}.
\label{eq:infornce}
\end{aligned}
\end{equation}
This equation is similar to the noise-contrastive estimation (NCE)~\cite{gutmann2010noise,he2020momentum} process. Minimizing~\cref{eq:cls_loss} is thereby tantamount to maximizing the lower bound of mutual information (MI) between the textual tag embeddings $\vc$ and the visual image embedding $\vz$.  More specifically, 
through the optimization of~\cref{eq:cls_loss}, the model is trained to align visual and text representations. This alignment is achieved by encouraging positive textual tag embeddings to be close to the visual image embedding and simultaneously ensuring that negative textual tag embeddings are farther away from the visual image embedding. As the image embedding represents a ``bag'' of patch embeddings in MIL, this alignment facilitates capturing the fine-grained semantic relationship between patches and tags.

\noindent\textbf{Discussion with GroupViT.}
GroupViT~\cite{xu2022groupvit} introduces a multi-label contrastive loss, which shares some similarities with our multi-tag classification loss. In GroupViT, nouns are extracted from captions using NLTK, and then handcrafted templates are used to prompt these nouns. These prompted texts are treated as positive samples during the contrastive learning process.
However, the multi-label contrastive loss in GroupVit primarily focuses on instance discrimination, where the prompted texts from different images are treated as negative samples. As a result, the semantic relationships between tags originating from different images are not explicitly considered. In contrast, our approach explicitly addresses the semantic relationships between tags from different images through a multi-tag classification loss.

\noindent\textbf{Handing Long-tailed Distribution.} As depicted in~\cref{fig:tags}(c), the distribution of the tags is highly imbalanced, even after filtering certain tags in the tail. To mitigate this issue,  we propose to utilize the balanced softmax strategy~\cite{ren2020balanced}. Specifically,  we adjust prediction logits by multiplying by the tag
frequencies. That is, we transform~\cref{eq:cls_score} into
\begin{equation}
\begin{aligned}
\vp_k = \frac{
\vw_k \cdot \exp(\rho \cdot \operatorname{cos}({\vz}, \vc_k) )
}
{\sum_{i=1}^{K} \vw_i \cdot \exp(\rho \cdot \operatorname{cos}({\vz}, \vc_i)).
}.
\label{eq:balanced_score} 
\end{aligned}
\end{equation}
where $\vw_k$ indicates the occurrence frequency of the $k$-th tag in the dataset.
In this way, the bias of imbalance can be alleviated by the frequency prior before computing final losses.

\subsection{Training and Inference} 

\noindent\textbf{Training.} In addition to the multi-tag classification loss, we can further enhance our method by incorporating the traditional image-text contrastive loss~\cite{radford2021learning}. Specifically, the image-text contrastive loss aims to encourage the alignment of features between the image and its corresponding caption, while simultaneously pushing away the features of other images and captions. 
Formally, the batch-wise image-text contrastive loss is expressed as:
\begin{equation}
\begin{aligned} 
    & \mathcal{L}^\mathrm{i2t}_\mathrm{Contrast}= - \sum_{n=1}^{N} \log \frac{
    \exp(\gamma \cdot \operatorname{cos}({\vz}_n, \vt_n) )
    }{\sum_{i=1}^{N} \exp(\gamma\cdot \operatorname{cos}({\vz}_n, \vt_i))
    }, 
    \\
    & \mathcal{L}^\mathrm{t2i}_\mathrm{Contrast}= - \sum_{n=1}^{N} \log \frac{
    \exp(\gamma \cdot \operatorname{cos}({\vt}_n, \vz_n) )
    }
    {\sum_{i=1}^{N} \exp(\gamma\cdot \operatorname{cos}({\vt}_n, \vz_i))
    }, 
    \\
    & \mathcal{L}_\mathrm{Contrast} = \mathcal{L}^\mathrm{i2t}_\mathrm{Contrast} + \mathcal{L}^\mathrm{t2i}_\mathrm{Contrast}. 
    \label{eq:contrastive_loss}
\end{aligned} 
\end{equation}
Here, with a slight abuse of notation, $N$ represents the number of image-text pairs within a batch, and ${\vz}_n$ and ${\vt}_n$ denote the feature of the $n$-th image and the feature of the $n$-th caption, respectively.    

Therefore, the overall training objective of our method is:
\begin{equation}
\begin{aligned} 
    \mathcal{L} = \mathcal{L}_\mathrm{Tag} + \lambda \mathcal{L}_\mathrm{Contrast}, 
    \label{eq:total_loss}
\end{aligned} 
\end{equation}
where $\lambda$ is the weight of the $\mathcal{L}_\mathrm{Contrast}$. We set $\lambda$ to $1$ in all our experiments.

\noindent\textbf{Inference.} 
We quantitatively evaluate our model over segmentation datasets to showcase its ability in achieving fine-grained image-text alignment.
For generating the segmentation results, 
we begin by reshaping the patch feature map $\rmZ \in \mathbb{R}^{L \times D}$ to a spatially structured format $ \rmZ \in \mathbb{R}^{H \times W \times D}$, where $L = HW$.
Next, we compute the similarity map between the patch features and the class embeddings. The class embeddings are generated by applying the prompt template ``a photo of a \{class name\}'' to the class names. 
Afterward, we upsample the similarity map to match the original image size through linear interpolation. Finally, the $\operatorname{Argmax}$ function is applied on the similarity map along the class dimension, thereby assigning each pixel to a particular class. 

%% file: sections/4.exp.tex
\section{Experiments}\label{sec:exp}

\subsection{Datasets and Evaluation Metric}
In the training phase, we use Conceptual 12M (CC12M) \cite{sharma2018conceptual} as the training dataset, which contains 12M image-text pairs. 
To provide an extensive evaluation, 
we evaluate our method on $10$ semantic segmentation benchmark datasets: PASCAL VOC~\cite{everingham2010pascal} (21 classes), PASCAL Context~\cite{mottaghi2014role} (60 classes), COCO-Object~\cite{caesar2018coco} (81 classes), Imagenet-S50~\cite{gao2022large} (50 classes), ImageNet-S300~\cite{gao2022large} (300 classes),  PASCAL VOC20~\cite{everingham2010pascal} (20 classes), PASCAL Context59~\cite{mottaghi2014role} (59 classes), COCO-Stuff~\cite{caesar2018coco} (171 classes), Cityscapes~\cite{cordts2016cityscapes} (19 classes), and ADE20K~\cite{zhou2019semantic} (150 classes), and $3$ referring image segmentation benchmark datasets: RefCOCO~\cite{yu2016modeling}, RefCOCO+~\cite{yu2016modeling}, and RefCOCOg~\cite{mao2016generation}. 

We employ mean intersection-over-union (mIoU) to evaluate the performance, which is a standard metric in semantic segmentation. 

\subsection{Implementation Details}
In our implementation, we utilize the Vicuna~\cite{chiang2023vicuna} as the large language model (LLM) for extracting tags. The visual encoder is based on the CLIP ViT-B/16 model. The input images are resized to a resolution of $224 \times 224$ pixels, and the patch size used is $16 \times 16$. We follow the modifications described in~\cite{li2023clip} to adapt the CLIP image encoder for our method.
For the projector, we adopt the design proposed in~\cite{cha2023learning}, which consists of two gated convolution blocks, but we exclude the upsampling interpolations in our implementation. In our framework, only the lightweight projector is trainable, while all other modules are fixed.
All our experiments are conducted on a single node equipped with 8 NVIDIA A100 GPUs, with each GPU having 80GB of memory. We train the model with a batch size of $4,096$ and a constant learning rate of \emph{1e-3} for a total of $30, 000$ iterations. Notably, the learning rate in our approach is not adjusted using any heuristic tricks (\eg, warmup or decay schedule), thereby reducing the number of hyperparameters involved in the training process. 
During inference, following the CLIP-based method TCL~\cite{cha2023learning}, we combine the segmentation results of our method and those of original CLIP through a weighted sum. Please refer to the \textbf{Appendix} for more details.

\begin{table*}[t]
    \centering
    \caption{ 
    \textbf{Zero-shot segmentation performance comparison on 10 object-related semantic segmentation datasets.}
    mIoU (\%) metric is used in every experiment. `Post' indicates whether any post-processing tricks (\eg, CRF~\cite{yi2023simple} or PAMR~\cite{cha2023learning,xing2023rewrite}) are employed during the inference stage.  `CC12M*' denotes that the OVSegmentor constructs the CC4M dataset by filtering CC12M with frequently appeared entities. Each dataset abbreviation stands for VOC: PASCAL VOC, Context: PASCAL Context, Object: COCO-Object, IN: ImageNet-S, Stuff: COCO-Stuff, City: Cityscapes, ADE: ADE20K.  
    }
    \renewcommand{\arraystretch}{1.0}
    \resizebox{\textwidth}{!}{
    \begin{tabular}{l|c|l|>{\columncolor[RGB]{230, 236, 227}}c>{\columncolor[RGB]{230, 236, 227}}c>{\columncolor[RGB]{230, 236, 227}}c>{\columncolor[RGB]{230, 236, 227}}c>{\columncolor[RGB]{230, 236, 227}}c>{\columncolor[RGB]{255, 239, 224}}c|>{\columncolor[RGB]{230, 236, 227}}c>{\columncolor[RGB]{230, 236, 227}}c>{\columncolor[RGB]{230, 236, 227}}c>{\columncolor[RGB]{230, 236, 227}}c>{\columncolor[RGB]{230, 236, 227}}c>{\columncolor[RGB]{255, 239, 224}}c}
    \Xhline{1.2pt}
        \multirow{2}{*}{\textbf{Methods}} & \multirow{2}{*}{\textbf{Post}} & \multirow{2}{*}{\textbf{Datasets}} & \multicolumn{6}{c|}{with background class}  & \multicolumn{6}{c}{without background class}    \\
        & & & \cellcolor{white}VOC  &  \cellcolor{white}Context &  \cellcolor{white}Object   &  \cellcolor{white}IN50&  \cellcolor{white}IN300&   \cellcolor{white}Avg.&\cellcolor{white}VOC20 &  \cellcolor{white}Context59 &  \cellcolor{white}Stuff &  \cellcolor{white}City &  \cellcolor{white}ADE  &  \cellcolor{white}Avg. \\ 
        \hline
        CLIPpy~\cite{ranasinghe2023perceptual} & \ding{55} & HQITP-134M & 52.2 & -   & 32.0 & -  &-    &-  &-  &-  &-  &-  &-  &-     \\
        ZeroSeg~\cite{chen2023exploring}       & \ding{55} & CC12M+COCO & 40.8 & 20.4 & 20.2 & -    & -    &-& -    & -    & - & - & -  &   - \\
         ViL-Seg~\cite{liu2022open}        & \ding{55} & CC12M           & 37.3 & 18.9 & 18.1 & -    & -    & -    & -   & -   & -    & -    & -    & -\\
         ViewCo~\cite{ren2022viewco}         & \ding{55} & CC12M         & 45.7 & 20.8 & 20.6 & -    & -    & -    & -   & -   & -    & -    & -   & - \\
         MixReorg~\cite{cai2023mixreorg}       & \ding{55} & CC12M       & 47.9 & 23.9 & 21.3 & -    & -    & -    & -   & -   & -    & -    & -  &  -    \\
         PGSeg~\cite{zhang2023uncovering}        & \ding{55} & CC12M   & 49.0 & 20.6 & 22.9 & -    & -    & -    & -   & -   & -    & -    & -  &   -   \\
        ViewCo~\cite{ren2022viewco}         & \ding{55} & CC12M+YFCC    & 52.4 & 23.0 & 23.5 & - & -   & -    && -    & -    & -    & -   & \\
        SimSeg~\cite{yi2023simple}        & \ding{55} & CC3\&12M       &  53.8 & 23.5 & 25.7 & -  &-    & -   & - &- &- &- &- &- \\
         
        ReCo~\cite{shin2022reco}& \ding{55} & ImageNet1K               & 25.1 & 19.9 & 15.7 & -    & -    & -    & 57.7 & 22.3 & 14.8 & 21.1 & 11.2 & 25.4 \\
        Dong\etal~\cite{dong2023maskclip}   & \ding{55} & LAION-20M  & 38.8 & 23.6 & 20.6 & 25.9 & 11.7 & 18.1 & 74.9 & 26.4 & 16.4 & 12.6 &  9.8 & 28.0\\
        
        OVSegmentor~\cite{xu2023learning}       & \ding{55} & CC12M*   & 53.8 & 20.4 & 25.1 & 38.9 & 14.4 & 30.5 & 66.1  &  19.2  &   10.5 & 6.4 &  5.3 & \\

        GroupViT~\cite{xu2022groupvit}     & \ding{55} & CC12M+YFCC   & 49.5 & 19.0 & 24.3 & 44.3 & 23.5 & 32.1 & 74.1 & 20.8 & 12.6 &  6.9 &  8.7 & 24.6\\
        
        CoCu~\cite{xing2023rewrite}          & \ding{55} & CC3\&12M+YFCC & 49.7 & 22.8 & 22.0 & 46.7 & 24.7 & 33.2 & -   & - & 14.9 & 21.9 & 12.0 & \\
        
        SegCLIP~\cite{luo2023segclip}       & \ding{55} & CC12M+COCO   & 52.6 & 24.7 & 26.5 & 47.4 &  24.0 & 35.0 & 77.6 &24.8 & 16.1& 11.2& 8.8 & 27.7\\
    
        TCL~\cite{cha2023learning}          & \ding{55} & CC3\&12M     & 51.6 & 24.3 & 30.4 & 46.9 & 26.6& 35.8 &  77.5 & 30.3 & 19.6 & 23.1 & 14.9 & 33.1\\
        \hline
          TagAlign & \ding{55} & CC12M   
          &\bf{55.3 } & \bf{32.7}  &  \bf{35.1} & \bf{52.8} & \bf{34.4} & \bf42.1{\textcolor{red}{+6.3}} &  \bf 83.8 & \bf 35.1 & \bf 23.1 &  \bf 27.3 & \bf 16.9 & \bf 37.2{\textcolor{red}{+4.1}} \\
          
          \hline
        MixReorg~\cite{cai2023mixreorg}      & \ding{52}& CC12M        & 50.5 & 25.4 & 23.6 & -& -& -& -    & -    & -    & -    & -    & - \\
        ReCo~\cite{shin2022reco}           & \ding{52} & ImageNet1K     & 27.2 & 21.9 & 17.3 & -& -& - & 62.4 & 24.7 & 16.3 & 22.8 & 12.4& 27.7 \\ 
        Dong\etal~\cite{dong2023maskclip}      & \ding{52}& LAION-20M   & 37.2 & 22.6 & 18.9 & -& -& - & 72.1 & 25.3 & 15.1 & 11.2 &  9.0& 26.5 \\ 
        GroupViT~\cite{xu2022groupvit}       & \ding{52}& CC12M+YFCC  & 51.1 & 19.0 & 27.9 & -& -& - & 81.5 & 23.8 & 15.4 & 11.6 &  9.4& 28.3\\ 
        SimSeg~\cite{yi2023simple}         & \ding{52}& CC3\&12M       &  57.4 & 26.2 & 29.7 & -  & & -  &  -    & -    & -    & -    & -  &  -  \\
        CoCu~\cite{xing2023rewrite}         & \ding{52} & CC3\&12M+YFCC & 51.4 & 23.6 & 22.7 & 48.8 & 25.5& 34.4 & -    & -    & 15.2 & 22.1 & 12.3  &- \\
        TCL~\cite{cha2023learning}            & \ding{52}& CC3\&12M    & 55.0 & 30.4 & 31.6 & 50.0 & 29.9 & 39.4  & 83.2 & 33.9 & 22.4 & 24.0 & 17.1 & 36.1 \\
        
        \hline

        TagAlign                               & \ding{52}& CC12M          &   \bf 57.9  & \bf 34.3    &   \bf 35.4   & \bf 55.1 & \bf 35.1  & \bf 43.6{\textcolor{red}{+4.2}}& \bf 88.4 & \bf 38.6   & \bf 25.8   &  \bf 27.6    & \bf 17.3 & \bf 39.5{\textcolor{red}{+3.4}} \\
        \Xhline{1.2pt}
    \end{tabular}%
    }
\label{table:main}
\end{table*}

\subsection{Main Results}

\noindent\textbf{Zero-shot semantic segmentation.}
In~\Cref{table:main}, we summarize the zero-shot semantic segmentation performance comparison between our method and previous methods across 10 benchmark datasets, including datasets with and without a background class.  To ensure a fair comparison, we report the performance of previous methods with and without the use of post-processing techniques, such as CRF~\cite{yi2023simple} or PAMR~\cite{cha2023learning,xing2023rewrite}. Our experimental results demonstrate that our method achieves significant improvements over previous state-of-the-art (SOTA) methods, using only the CC12M dataset for training. For instance, without post-processing techniques, our method outperforms the previous SOTA method TCL~\cite{cha2023learning} by an average of $6.3\%$ and $4.1\%$ on datasets with and without a background class, respectively.
When post-processing techniques are applied, our method maintains its superiority over previous methods by $4.2\%$ and $3.4\%$. These results serve as compelling evidence for the effectiveness of our approach in zero-shot semantic segmentation.

\begin{table}[t!]
    \centering
    \scriptsize
    \SetTblrInner{rowsep=1.5pt}       
    \SetTblrInner{colsep=2.0pt}       
    \captionof{table}{
    \textbf{Zero-shot segmentation performance comparison on 3 referring segmentation datasets.} mIoU (\%) metric is used in every experiment.
    }
    \resizebox{\linewidth}{!}{
    \begin{tblr}{
        cells={halign=c,valign=m},  
        cell{1}{1}={halign=c},      
        hline{3}={1-9}{},   
        hline{2}={2-9}{},   
        hline{1,7}={1.0pt},         
        vline{2,5,8}={1-6}{},           
        cell{1}{1}={r=2}{},      
        cell{1}{2}={c=3}{},      
        cell{1}{5}={c=3}{},      
    }
    \textbf{Methods} & {RefCOCO} &&& {RefCOCO+}&& & {RefCOCOg}\\
    &  \textit{val} & \textit{testA} & \textit{testB}   & \textit{val} & \textit{testA} & \textit{testB}  & \textit{val}  \\ 
    GroupViT~\cite{xu2022groupvit}       &  7.99 &  6.16 & 10.51 &  8.49 &  6.79 & 10.59 & 10.68 \\
    MaskCLIP~\cite{zhou2022extract}       & 11.52 & 11.85 & 12.06 & 11.87 & 12.01 & 12.57 & 12.74 \\
    \hline
    TagAlign-Obj.   & 15.87& 17.52& 15.69& 16.04& 17.26& 16.44& 18.59  \\
    + Attri.     & \bf18.75& \bf20.31& \bf20.64& \bf19.24& \bf20.88&\bf21.23 &\bf23.69 \\ 
    \end{tblr}
    }
    \label{tab:comp_open}
    \vspace{-3mm}
\end{table}

\noindent\textbf{Zero-shot referring segmentation.}
Additionally, we evaluate the performance of our method on referring segmentation datasets. Unlike the aforementioned semantic segmentation task, which primarily focuses on segmenting a set of object classes, referring segmentation entails the segmentation of entities described in unstructured natural language text, which frequently encompasses objects with distinct attributes. The experimental results, as depicted in~\Cref{tab:comp_open}, clearly demonstrate the substantial performance improvement achieved by our method when attributes are incorporated, compared to the case where attributes are not utilized. These results validate the effectiveness and importance of attribute-guided learning in enhancing the performance of referring segmentation.

\subsection{Ablation Study}
In this section, we perform a series of ablation experiments to assess the individual impact of each technique employed in our method. In addition to the commonly used PASCAL VOC (VOC) dataset, we also evaluate our method on the ImageNet-S50 (IN50) and ImageNet-300 (IN300) datasets, providing insights into how our method performs across datasets that vary in the number of classes

\begin{table}[t!]
    \centering\scriptsize
    \SetTblrInner{rowsep=1.5pt}       
    \SetTblrInner{colsep=7.0pt}       
    \captionof{table}{%
    \textbf{Ablation study on the number of object tag (\ie, $K$).} We report the performance where only the multi-tag classification loss $\mathcal{L}_\mathrm{Tag}$ is used.
    }
    \label{tab:ablation_k}
    \resizebox{\linewidth}{!}{
    \begin{tblr}{
        cells={halign=c,valign=m},  
        column{1}={halign=l},       
        hline{1,5}={1.0pt},        
        hline{2,3,4}={1-11}{},       
        vline{2}={1-8}{},       
    }
    $K$ & 2.5k & 7.5k & 10k & 25k & 50k & 100k \\
    VOC& 45.1&45.0&45.7&45.1&45.0&44.8   \\
    IN50 & 27.2 & 28.8 & 29.2 & 28.9 & 29.3 & 29.4 \\
    IN300 & 15.0 & 17.4 & 17.8 & 18.0 & 18.8 & 19.0 \\
    \end{tblr}
    }
    \vspace{-3mm}
\end{table}

\noindent\textbf{Impact of the number of object tags.}
In this work, we establish the tag list by selecting the most frequent $K$ tags. As shown in~\Cref{tab:ablation_k}, we evaluate the performance for different values of $K$ by considering only the multi-tag classification loss $\mathcal{L}_\mathrm{Tag}$ and have the following observations. The performance on the ImageNet-S300 dataset, which contains more classes, tends to improve with larger values of $K$. This suggests that including a larger number of tags in the tag list allows the model to capture a wider range of object classes, leading to better performance on datasets with more diverse classes.
Conversely, on the ImageNet-S50 dataset, which contains fewer classes, the performance tends to be better with smaller values of $K$. This indicates that a smaller tag list, focused on a subset of the most frequent tags, is sufficient to cover the object classes present in datasets with fewer classes.
Considering the trade-off between complexity and performance, we choose $K=10,000$ as the default value in our work. Increasing $K$ beyond this value leads to marginal improvements.

\begin{table}[t!]
    \centering\scriptsize
    \SetTblrInner{rowsep=1.5pt}       
    \SetTblrInner{colsep=6.9pt}       
    \captionof{table}{%
         \textbf{Ablation study on the tag parsing methods,} including Tag2text, NLTK, and LLM. ``No. of Tags'' indicates the total number of object tags and attribute tags. 
    }
    \label{tab:tags}
    \resizebox{\linewidth}{!}{
    \begin{tblr}{
        cells={halign=c,valign=m},  
        column{1}={halign=c},       
        hline{1,2,3,4}={1-5}{},  
        hline{1,5}={1.0pt},        
    }
        Text Parser & No. of Tags & VOC& IN50 & IN300   \\
        LLM         & 20000 & \bf55.3 &   \bf52.8  & \bf34.4   \\
        NLTK        & 20000 & 51.0 &   49.1 & 31.2  \\
        Tag2text~\cite{huang2023tag2text}        & 4585& 50.1 &  48.2  & 32.1 \\
    \end{tblr}
    }
\end{table}

\noindent\textbf{Impact of text parsing methods.}
Here, we compare the performance of using NLTK and LLM for tag extraction, as depicted in the first two rows of~\Cref{tab:tags}. Our results demonstrate that LLM outperforms NLTK significantly, highlighting the superiority of our proposed LLM-aided parsing method.

Additionally, we refer to Tag2text~\cite{huang2023tag2text}, which also employs NLTK for tag extraction. However, Tag2text retains a limited set of $4,585$ tags encompassing both objects and attributes. To assess the impact of this restricted vocabulary, we conduct an experiment where we solely consider tags employed by Tag2text. The corresponding results are presented in the last row of~\Cref{tab:tags}. Notably, this approach performs worse compared to the other two methods. This outcome can be attributed to the limited open vocabulary capacity of the tags.

\begin{table}[t!]
    \centering\scriptsize
    \SetTblrInner{rowsep=1.5pt}       
    \SetTblrInner{colsep=7.5pt}       
    \captionof{table}{%
         \textbf{Performance comparison with DiHT~\cite{radenovic2023filtering}} that distills a pre-trained multi-tag classification vision model to CLIP.   
    }
    \label{tab:diht}
    \resizebox{\linewidth}{!}{
    \begin{tblr}{
        cells={halign=c,valign=m},  
        column{1}={halign=c},       
        hline{1,2,3,4}={1-5}{},  
        hline{1,4}={1.0pt},        
    }
        Method   &Dataset& VOC & IN50 & IN300  \\
        TagAlign &CC12M&  55.3 &   52.8  & 34.4      \\
        DiHT~\cite{radenovic2023filtering}     &Laion-2B&    13.5 &  10.6 &    9.7   \\
    \end{tblr}
    }
\end{table}

\noindent\textbf{Performance comparison with DiHT~\cite{radenovic2023filtering}.}
Here, we compare our TagAlign with another method that also utilizes multi-tag classification to enhance CLIP, \ie, DiHT~\cite{radenovic2023filtering}. The results of this comparison are presented in~\Cref{tab:diht}. From the table, it can be observed that DiHT performs poorly on the semantic segmentation benchmarks compared to our method. There are a couple of reasons for this performance difference. 1) DiHT does not explicitly align the visual embeddings and textual tag embeddings. 2) DiHT primarily concentrates on leveraging multi-tag classification to enhance the image-level representations.

\noindent\textbf{Impact of the long-tailed training loss.}
As illustrated in~\cref{fig:tags}(b), the extracted tags are highly imbalanced. Here we explore various loss functions aimed at mitigating tag imbalance, including balanced softmax loss~\cite{ren2020balanced}, weighted softmax loss~\cite{zhang2023deep}, focal loss~\cite{lin2017focal}, and asymmetric loss~\cite{ridnik2021asymmetric}. The experimental results in~\Cref{tab:quantitative_pose} demonstrate that the balanced softmax loss outperforms other alternatives, making it our default choice.

\noindent\textbf{Combining with the state-of-the-art methods in different architectures.}
In this study, we utilize the ViT of CLIP as the primary visual encoder. To assess the versatility of our method across different visual encoders, we conduct experiments by applying our method to a GroupViT-based encoder~\cite{xu2022groupvit}, which is another popular visual encoder in the literature. The recent method OVSegmentor~\cite{xu2023learning} is adopted as the baseline for the GroupViT-based encoder. The performances of using different kinds of visual encoders are presented in~\Cref{tab:visualen}. As can be seen,  our method demonstrates significant performance improvements on both kinds of encoders, highlighting the generalization ability of our approach.

\begin{table}[t!]
    \centering\scriptsize
    \SetTblrInner{rowsep=1.5pt}       
    \SetTblrInner{colsep=8.0pt}       
   \captionof{table}{%
        \textbf{Ablation study on the long-tailed training loss.}
    }
    \label{tab:quantitative_pose}
    \resizebox{\linewidth}{!}{
    \begin{tblr}{
        cells={halign=c,valign=m},  
        column{1}={halign=c},       
        hline{1,6}={1.0pt},        
        hline{2,3,4,5}={1-4}{},       
    }
        {Method}& VOC & IN50 & IN300  \\
        Balanced Softmax Loss & \bf55.3 &  \bf52.8  & \bf34.4 \\
        Weighted Softmax Loss & 52.9 & 49.4 & 32.7 \\
        Focal Loss & 52.0& 48.7 & 32.6  \\
        Asymmetric Loss & 52.7 & 50.0 & 32.3  \\
    \end{tblr}
    }
\end{table}

\begin{table}[t!]
    \centering\scriptsize
    \SetTblrInner{rowsep=1.5pt}       
    \SetTblrInner{colsep=9.0pt}       
    \captionof{table}{%
        \textbf{Combining with SOTA methods in different architectures}. 
    }
    \label{tab:visualen}
    \resizebox{\linewidth}{!}{
    \begin{tblr}{
        cells={halign=c,valign=m},  
        column{1}={halign=c},       
        hline{1,2,4,6}={1-6}{},
        hline{1,6}={1.0pt},        
    } 
        Method      & Architecture & IN50 & IN300  \\
        OVSegmentor & GroupViT  & 38.9 & 14.4   \\
        + Multi-Tags & GroupViT  & 41.0 & 15.7  \\
        CLIP        & ViT  & 47.2 & 27.1   \\
        + Multi-Tags & ViT  &  \bf52.8  & \bf34.4 \\
    \end{tblr}
    }
    \vspace{-3mm}
\end{table} 

\begin{figure*}[t]
	\centering
	\includegraphics[width=0.95\linewidth]{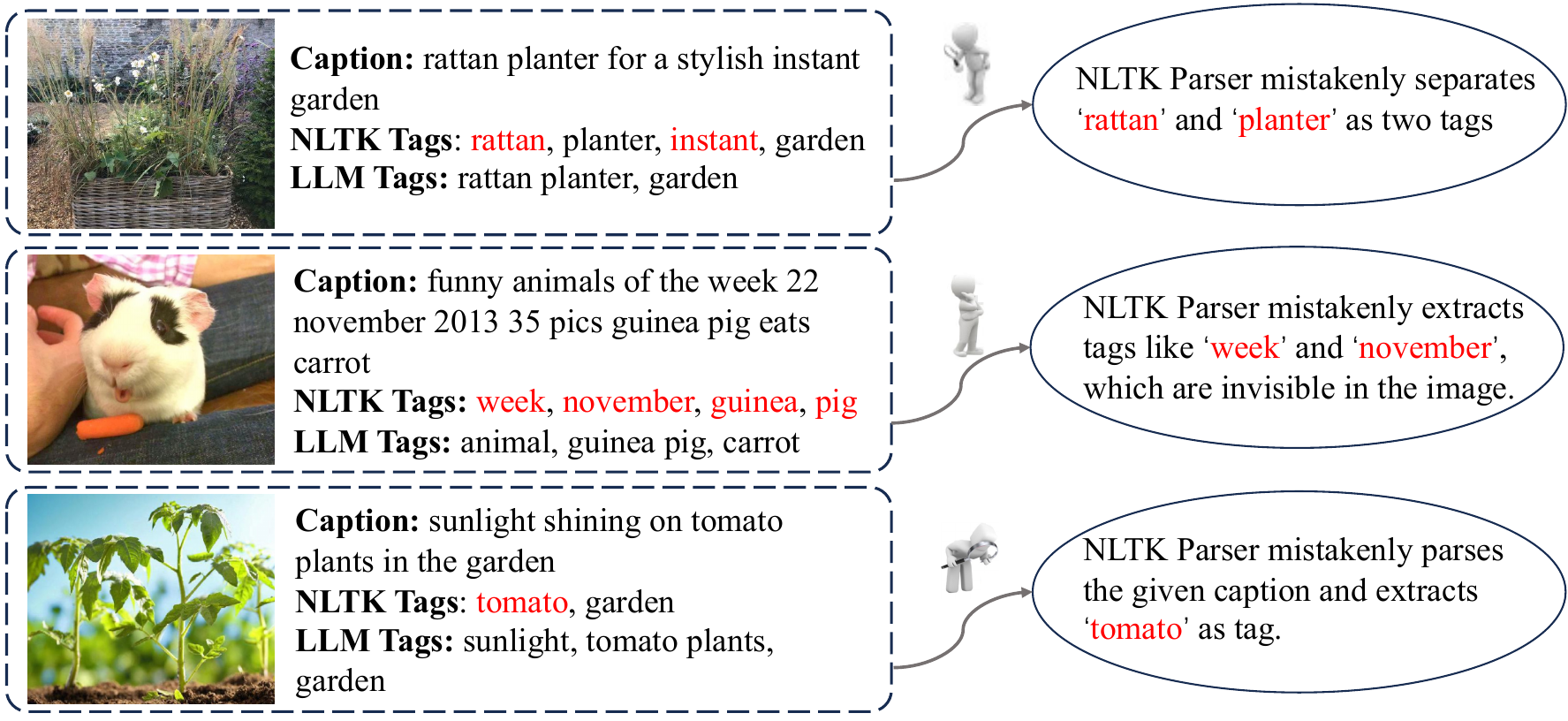}
	\centering
	\caption{\textbf{Comparison of tags extracted with text parsing methods (\textit{i.e.,} NLTK and LLM).} The tags in red are misidentified by the parser. Best viewed in color.
 } 
	\label{fig:compare_tags}
\end{figure*}

\begin{figure*}[t]
	\centering
	\includegraphics[width=\linewidth]{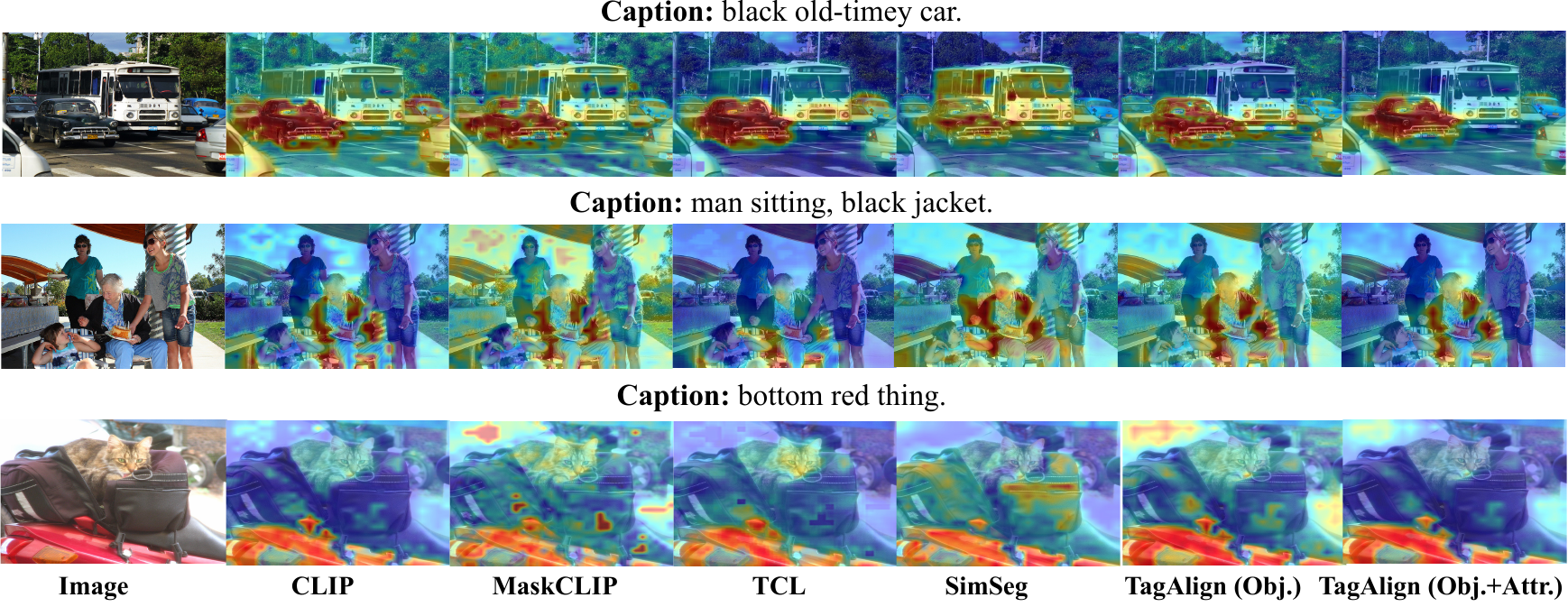}
	\centering
 \caption{
 \textbf{Visualization of the similarity maps} between patch features and text features generated by different methods. Best viewed in color.} \label{fig:compare_vis}
\end{figure*}

\subsection{Visualization}
In~\cref{fig:compare_vis}, we visualize the similarity maps between text features and image features obtained from CLIP, TagAlign (Obj.), TagAlign (Obj.+Attr.), and existing language-superivsed semantic segmentation methods, \ie MaskCLIP~\cite{ding2023maskclip}, TCL~\cite{cha2023learning}, and SimSeg~\cite{yi2023simple}. 
We notice that the background exhibits high activation when aligning text features with image features extracted by CLIP and existing methods.
The similarity maps obtained from TagAlign (Obj.) indicate the alignment between text and image features is strongly affected by object-related regions, \eg, ``car'' (first row) and ``man'' (second row).
By incorporating attribute tags, TagAlign (Obj.+Attr.) can identify a more appropriate region within the given image based on the description.
In addition, we visualize the similarity maps of previous methods including MaskCLIP~\cite{zhou2022extract}, TCL~\cite{cha2023learning} and SimSeg~\cite{yi2023simple}.
The comparison reveals that the qualitative results of TagAlign (Obj.+Attr.) surpass those obtained from the aforementioned methods.
The visualization results further prove the capacity of TagAlign to prompt more precise alignment between image-text encoders. 

%% file: sections/5.conclusion.tex
\section{Conclusion}\label{sec:conclusion}
In this work, we design a simple yet effective approach to assist precise alignment of the vision-language model without requiring extra annotations.
Specifically, we employ a large language model to automatically parse object tags and attribute tags from the caption. 
These parsed tags are utilized to supervise the training procedure with multi-tag classification loss, in cooperation with the commonly used image-text contrastive loss. 
Extensive experiments demonstrate that our method achieves remarkable performances on 10 semantic segmentation benchmarks and 3 referring expression segmentation benchmarks. Besides, visual examples illustrate that integrating object and attribute supervision significantly improves the localization of attribute-specified objects.  This inspires our future direction: delving into a wider variety of tag types (\eg, object relation) to further boost the vision-language alignment.

%% file: sections/6.ref.tex
\bibliographystyle{ieeenat_fullname}
\bibliography{ref.bib}

%% file: sections/7.appendix.tex
\clearpage
\appendix

\section{Pseudo Code} \label{A:pseudo_code} 
For clarity, we present the pseudo-code for the core implementation of TagAlign in~\cref{alg:code}. As can be seen, the pipeline of TagAlign is implementation-friendly. 

\begin{algorithm}[!h]
\caption{\small Pseudo code of TagAlign in Pytorch-like style.
	}
	\label{alg:code}
	\definecolor{codeblue}{rgb}{0.25,0.5,0.5}
	\definecolor{codekw}{rgb}{0.85, 0.18, 0.50}
	\lstset{
		backgroundcolor=\color{white},
		basicstyle=\fontsize{8pt}{8pt}\ttfamily\selectfont,
		columns=fullflexible,
		breaklines=true,
		captionpos=b,
		numbers=left,
		xleftmargin=1em,
		commentstyle=\fontsize{8pt}{8pt}\color{codeblue},
		keywordstyle=\fontsize{8pt}{8pt}\color{codekw},
		escapechar=\&
	}
	\begin{lstlisting}[language=python]
# x_v: a batch of images (B, H, W, 3)
# x_t: a batch of captions (B, L_t), where L_t is the length of the captions
# x_w: the prompted text of the tags (K, L_t)
# y: the binarized labels of the input images (B, K)
# image_enc: the visual encoder of CLIP
# text_enc: the text encoder of CLIP
# proj: the projector

# generate patch embedding
z_p = proj(image_enc(x_v)) # (B, D, L), where L is the number of patches
# apply global average pooling to generate image embedding
z = mean(z_p, dim=-1) # (B, D)
# generate caption embedding
t = text_enc(x_t) # (B, D)
# generate tag embedding
w = text_enc(x_w) # (K, D)

# compute tag classification probability
p = softmax(z @ w.t(), dim=-1) # (B, K)
# compute tag classification loss
l_tag = cross_entropy(p, y)

# compute image-text contrastive loss 
l_nce = infor_nce(z, t) + infor_nce(t, z)

l = l_tag + l_nce

\end{lstlisting}

\end{algorithm}

\section{Discussion with Related Works} \label{A:related_work}
In the main manuscript, we discussed the differences between our method and previous methods in the Related Work section. In this section, we would like to further highlight the differences between our method and the series works of RAM (\ie, Tag2text~\cite{huang2023tag2text} and RAM~\cite{ram}).
These works also enhance vision-language models by incorporating tag information extracted from image captions. However, these methods focus on classifying the tag while overlooking the fine-grained image-text alignment, which makes them fundamentally different from our method. 

Specifically, in both Tag2text~\cite{huang2023tag2text} and RAM~\cite{ram}, a decoder module is employed to classify commonly used tags. This module incorporates multiple cross-attention layers, where the tag features serve as queries and the patch features serve as keys/values. Through this mechanism, the decoder module produces updated tag embeddings that have interacted with the patch features. Subsequently, the tag embeddings are fed to a fully connected layer to predict the probabilities of the image belonging to different tags.
In contrast, our method aggregates patch features into an image feature through average pooling, and then~\textbf{explicitly} aligns the tag features with the image feature via contrastive loss (\ie, the multi-label classification loss). Additionally, based on the principle of multiple-instance learning, this way also explicitly aligns the tag features with the patch features.

Hence, compared with Tag2text~\cite{huang2023tag2text} and RAM~\cite{ram}, our method offers two distinct advantages:
1) Fine-grained alignment between patch features and text features: By explicitly aligning the tag features with both the image feature representation and the individual patch features, TagAlign achieves a fine-grained alignment between visual and textual elements. This fine-grained alignment enables TagAlign to perform tasks that require detailed visual-textual correspondence, \eg, semantic segmentation. In contrast, it is challenging to adapt Tag2text and RAM for such fine-grained tasks. 
2) Flexibility and robustness in recognizing a wider range of open-world concepts: TagAlign computes the similarity between visual features and text features using cosine similarity, without relying on a heavy decoder module as in Tag2text and RAM. This approach allows for greater flexibility in recognizing diverse concepts and leveraging the capabilities of text embeddings from CLIP text encoders. Notably, CLIP itself is trained using cosine similarity between visual features and text features, making TagAlign particularly effective in leveraging the strengths of CLIP. The experimental section of the manuscript demonstrates that the tags defined in Tag2text are limited in their generalizability on open-world datasets.

\section{Implementation Details} \label{A:imple}  
\subsection{Architecture Details}
\paragraph{Large language model.} We employ the Vicuna-33b~\cite{chiang2023vicuna} as our large language model (LLM) to extract tags. The prompts we have devised are depicted in~\Cref{tab:tag_extract_prompt}.
It is important to note that the tag parsing process is carried out offline and performed only once, rendering it cost-effective in practical scenarios.  

\paragraph{Visual encoder.} 
The visual encoder is based on the CLIP ViT-B/16 model~\cite{radford2021learning}, while we follow the modifications described in~\cite{li2023clip} to adapt the CLIP visual encoder for our method.
Specifically, we maintain the existing architecture in the shallow layers while introducing changes in the deep layers, where we replace the k-q attention operations with v-v attention operations and remove the feed-forward network (FFN) within these deep layers. Please refer to~\cite{li2023clip} for more details about the encoder structure. 

\paragraph{Projector.}  Regarding the projector, we adopt the structure of the decoder proposed in~\cite{xu2023learning}. This projector comprises two gated convolution blocks. However, we deviate from the original implementation by removing the upsampling interpolations that were utilized in~\cite{xu2023learning}. We refer to~\cite{xu2023learning} for more details about the structure of the gated convolution blocks. 

\begin{table}[!h]
\centering
\caption{\textbf{Prompts to extract tags from the caption.}
For each query, we illustrate the prompt construction process for Vicuna to collect the conversation response.}
\label{tab:prompt_conversation}
\begin{minipage}{0.99\columnwidth}\vspace{0mm}    \centering
\begin{tcolorbox} 
    \centering
    \small
    \begin{tabular}{p{0.99\columnwidth}}

\begin{minipage}{0.99\columnwidth}

\VarSty{messages} = [
%
\{\var{"role":"system", "content":} f"""You are a Natural Language Processing (NLP) expert. I will provide you with a caption that describes a image.
Please analyze the caption and identify all distinct physical objects mentioned and specific attributes. \\ 
\\
When you identify objects, please follow these guidelines:   \\  %
(1) Analyze the caption and identify objects specifically mentioned.\\
(2) If an object is a proper noun or specified by a brand or model, convert it to its generic noun form.\\
(3) Only list tangible and visible objects that would likely be the primary focus or clear elements within an image, excluding abstract concepts or items inferred but not directly stated.\\
(4) Exclude any adjectives that may describe a noun. Only the primary object nouns should be included in your response.\\
\\
Based on the identified objects, please analyze the specific attribute of them following these guidelines:\\
(5) Analyze the attribute of the identified objects specifically mentioned.\\
(6) If one attribute consists of multiple words, shorten it to one word.\\ 
\\
I will provide the caption. Your response should be in the format of ``objects: [object1, object2, ...], attributes:[(attribute1, object1), (attribute2, object2),...]''. If there is no attribute in the caption, Your response should be in the format of ``objects: [object1, object2, ...], attributes:[]''. If the caption, for example, reads ``An image of two grey Maine Coon cats sitting next to a blue IKEA bowl filled with daisies on a wooden Crate \&  Barrel table with a Sting Ray parked in the background'', you should reply with ``objects: [cat, bowl, daisy, table, car], attributes: [(grey, cat), (two, cat), (blue, bowl), (wooden, table)]''. """\}]

\end{minipage}
    \end{tabular}
\end{tcolorbox}
\end{minipage}
\label{tab:tag_extract_prompt}
\end{table}

\subsection{Training Details}
Our method is trained on a single node equipped with $8$ NVIDIA A100 GPUs, with each GPU having $80$GB of memory. We train the model with a batch size of $4,096$ and a constant learning rate of \emph{1e-3} for a total of $30, 000$ iterations. We use AdamW as the optimizer. During training, only the parameters of the projector are trainable. The model takes less than $12$ hours to train in total.

\subsection{Testing Details}\label{sec:testing}
As described in the main paper, our approach involves computing the cosine similarity map between patch features and text features of different classes. 
Subsequently, the similarity map is upsampled to match the shape of the input image. We denote the upsampled similarity map as $\rmM \in \mathbb{R}^{H\times W \times C}$, where $H$ and $W$ represent the spatial dimensions of the image, and $C$ is the number of classes in the testing dataset.

In our implementation, we calculate two similarity maps, referred to as $\rmM_{\text{TagAlign}}$ and $\rmM_{\text{CLIP}}$. The first one, $\rmM_{\text{TagAlign}}$, uses the patch features after the projector, \ie, the features produced by our TagAlign model. The second one, $\rmM_{\text{CLIP}}$, is based on the patch features before the projector, \ie, the features produced by the original CLIP model. Following~\cite{cha2023learning}, we merge $\rmM_{\text{TagAlign}}$ and $\rmM_{\text{CLIP}}$ using a weighted sum to obtain the final similarity map $\rmM$:
\begin{equation}
\begin{aligned}
\rmM = \alpha \rmM_{\text{TagAlign}} + (1 - \alpha) \rmM_{\text{CLIP}}
\label{eq:weighted_sum}
\end{aligned}
\end{equation}
where $\alpha$ is the pre-defined weight.

Afterward, we apply $\operatorname{Softmax}$ over the similarity map $\rmM$ along the class dimension:
\begin{equation}
\begin{aligned}
\hat{\rmM} = \operatorname{Softmax}(s \rmM),
\label{eq:mask}
\end{aligned}
\end{equation}
where $s$ is a scaling factor.
$\hat{\rmM}_{i,j} \in \mathbb{R}^{C}$ denotes the probabilities of each pixel belonging to the $C$ classes. The $\operatorname{Argmax}$ opration is applied over $\hat{\rmM}$ for assigning each pixel to a specific class. In scenarios where the testing dataset includes a background class, we apply a thresholding strategy to $\hat{\rmM}$, where a pixel is assigned to the background class if the maximum value of $\hat{\rmM}_{i,j}$ is smaller than a certain threshold.   

Following common practice~\cite{xu2022groupvit}, we resize each input image to have a shorter side of size $448$ pixels during inference. In addition, sliding windows are cropped from the image with a fixed stride and the segmentation results of different sliding windows are fused together.

As stated above, there are two inference techniques used to enhance the performance: 1) combining with CLIP; 2) using sliding windows over the image.  The experiments detailed in Table~\ref{tab:ablation_inference_techniques} are designed to assess the impact of these techniques. Additionally, the previous best performance of other methods is listed in the second row for comparison.
As can be seen, while these techniques do enhance our method's performance, they are not the primary factors contributing to superiority over other methods. 

\begin{table}[!h]
    \centering
    \scriptsize
    \SetTblrInner{rowsep=2.0pt}       
    \SetTblrInner{colsep=8.0pt}       
    \captionof{table}{%
    \textbf{Ablation study on the inference techniques.}   
    }
    \label{tab:ablation_inference_techniques}
    \begin{tblr}{
        cells={halign=c,valign=m},  
        column{1}={halign=l},       
        hline{1,6}={1.0pt},        
        hline{2,3}={1-4}{},       
        vline{2}={1-5}{},       
    }
        Method & VOC & IN50 & IN300   \\
        second competitor & 53.8 & 47.4 & 26.6 \\
        plain TagAlign &  54.5 & 51.1 & 33.0 \\
        + combine with CLIP & 54.5 & 51.6 & 33.2   \\
        + sliding windows $\rightarrow$ TagAlign & 55.3 & 52.8 & 34.4  \\
    \end{tblr}
\end{table}

\section{Visualization Examples} \label{A:vis}
\begin{figure*}[!h]
	\centering
        \includegraphics[width=0.8\linewidth]{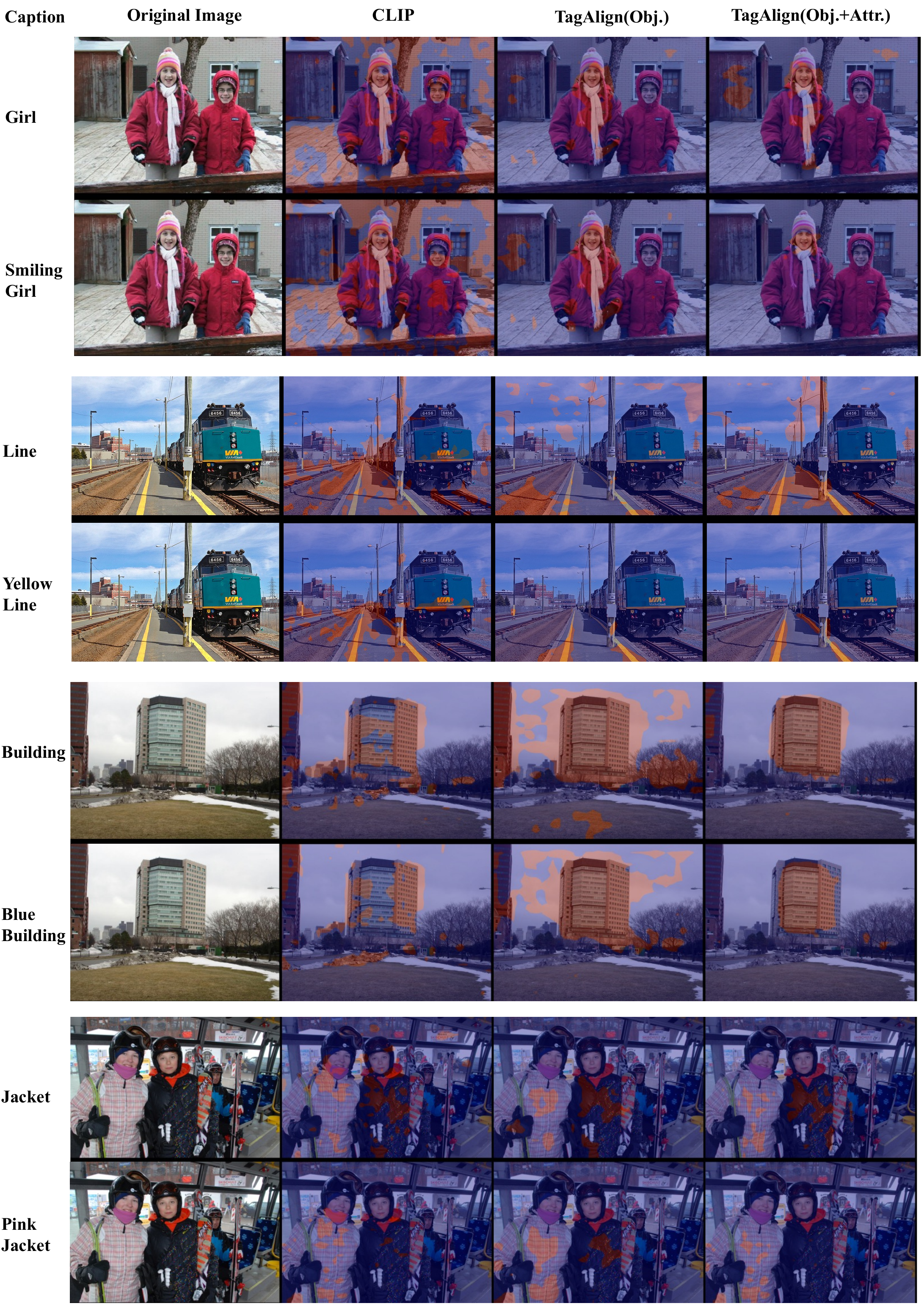}
	\centering
	\caption{
        \textbf{Visualized examples for different methods.} ``CLIP'' denotes the original CLIP model. ``TagAlign(Obj.)'' indicates the utilization of object tags, while ``TagAlign(Obj.+Attr.)'' incorporates both object tags and attribute tags.
        } 
	\label{fig:vis_supple}
\end{figure*}

In addition to the visualized examples provided in the main manuscript, we present more examples in~\cref{fig:vis_supple} to further reinforce the reliability of our observations. These additional examples demonstrate consistent findings that align with the observations made in the main manuscript.

Specifically, we find that TagAlign (Obj.) and TagAlign (Obj.+Attr.) outperform CLIP in terms of segmenting noun-based objects (\eg, ``Girl'', ``Line''). Furthermore, we observe that TagAlign (Obj.+Attr.) exhibits superior performance compared to TagAlign (Obj.) in segmenting objects specified by the attributes (\eg, ``Smiling Girl'', 
 ``Yellow Line’‘).

\begin{figure*}[!h]
	\centering
        \includegraphics[width=0.8\linewidth]{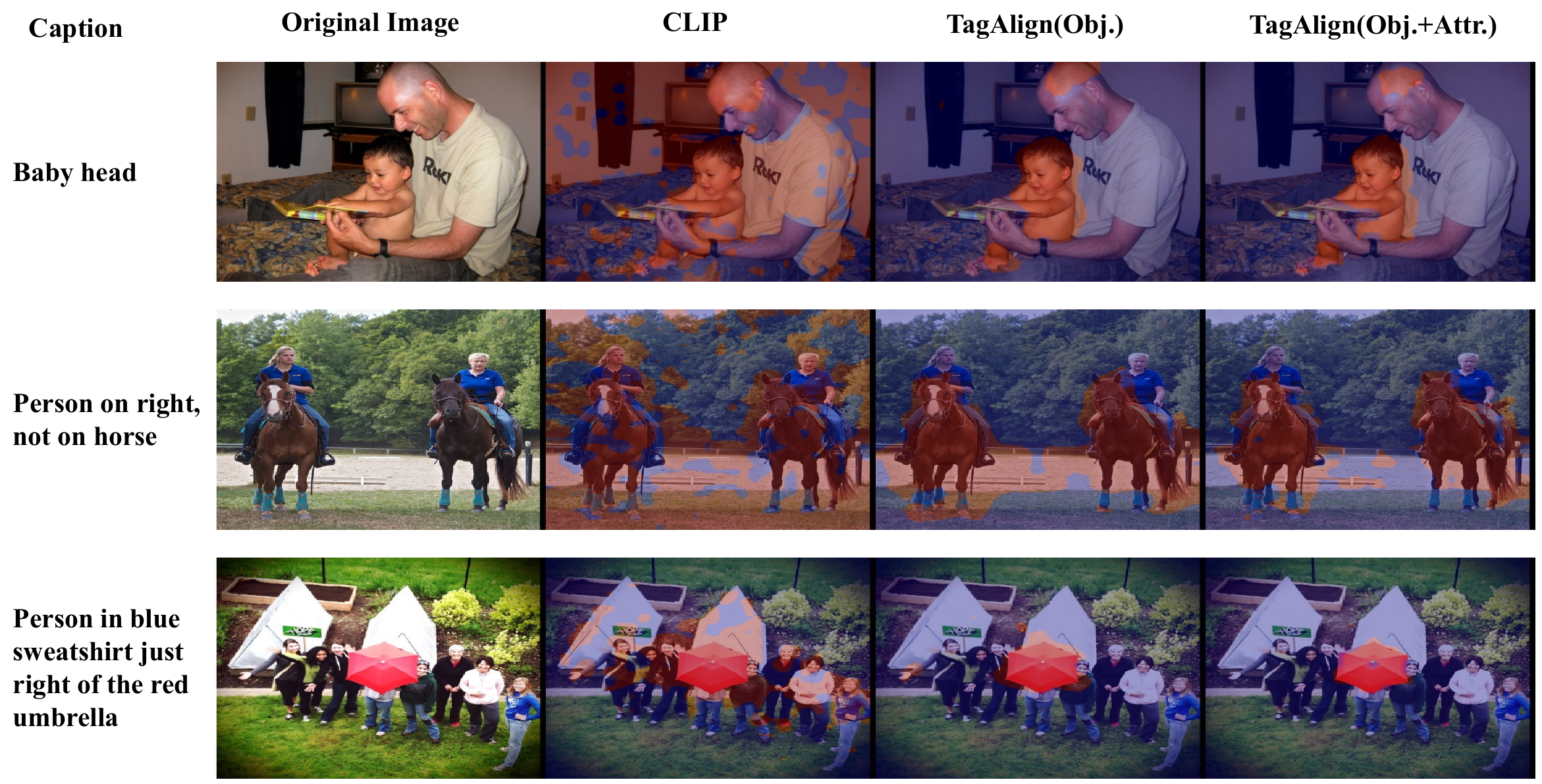}
	\centering
	\caption{
        \textbf{Samples of failure cases.} The presented examples illustrate instances where both CLIP and TagAlign encounter challenges in accurately segmenting the given queries (captions).
        } 
	\label{fig:fail_case}
\end{figure*}
These additional visualized examples provide further support for our observations, thereby reinforcing the effectiveness of TagAlign (Obj.) and TagAlign (Obj.+Attr.) in accurately aligning text features and image features compared to CLIP.

\section{Limitation and Future Work} \label{A:failure} 
In this section, we present several examples that demonstrate failure cases of our method, as depicted in~\cref{fig:fail_case}. Based on our analysis, we attribute these failures to two main reasons:
1) Inaccurate understanding of the text: In certain cases, the model exhibits difficulties in accurately comprehending the textual queries. For instance, when presented with the query ``Baby Head'', the model tends to interpret both ``Baby'' and "Head" as separate queries, resulting in the highlighting of both the regions corresponding to ``Baby'' and ``Head''. Moreover, the model struggles to understand queries such as ''Person on right, not on horse'', leading to the simultaneous highlighting of both the ``person'' and the ``horse''.
2) Inaccurate understanding of object relations: The model faces challenges in comprehending complex object relations, as exemplified by its failure to correctly understand queries like ``Person in blue sweatshirt just right of the red umbrella.''

Based on the identified failure cases, we outline potential directions for future research:
1) Integration with a powerful large language model: We plan to connect our visual encoder with a large language model that possesses a better understanding of textual input. This integration can potentially enhance the model's ability to accurately comprehend and interpret complex queries.
2) Modeling object relations: In addition to capturing objects and attributes, we intend to explore the modeling of relationships between objects. By incorporating explicit modeling of object relations, we expect to improve the ability of the model to understand and reason about complex spatial and semantic relationships among objects.